\documentclass[10pt,twocolumn,letterpaper]{article}

\usepackage[pagenumbers]{wacv} 

\definecolor{wacvblue}{rgb}{0.21,0.49,0.74}
\usepackage[pagebackref,breaklinks,colorlinks,allcolors=wacvblue]{hyperref}

\usepackage{multirow}
\usepackage{flushend}
\usepackage{graphicx}
\usepackage{soul}
\usepackage{booktabs}
\usepackage{arydshln} 
\usepackage{colortbl}
\usepackage{amsmath, amssymb}
\usepackage{booktabs}
\usepackage{array}
\usepackage{comment}
\usepackage{adjustbox}
\usepackage{arydshln}
\usepackage{wrapfig}
\usepackage{colortbl}
\usepackage{subcaption} 
\usepackage{caption}
\usepackage{algorithm}
\usepackage{algpseudocode}
\usepackage{dblfloatfix}
\usepackage{fancyvrb}
\usepackage{alltt}
\usepackage{fancybox}
\usepackage[accsupp]{axessibility}

\usepackage{xcolor}
\usepackage{framed}
\usepackage{ragged2e}

\definecolor{cvprblue}{rgb}{0.21,0.49,0.74}
\definecolor{lightgray}{gray}{0.9}
\definecolor{darkblue}{RGB}{94,110,186}
\definecolor{darkGreen}{RGB}{92, 148, 110}
\definecolor{darkgreen}{RGB}{0,100,0}
\definecolor{lightorange}{RGB}{255,160,0}
\definecolor{darkred}{RGB}{139,0,0}

\definecolor{PastaYellow}{RGB}{255,230,190}
\definecolor{GroupColor1}{RGB}{220,230,255}
\definecolor{GroupColor2}{RGB}{220,255,220}
\definecolor{GroupColor3}{RGB}{255,220,200}
\definecolor{GroupColor4}{RGB}{230,220,255}
\definecolor{GroupColor5}{RGB}{210,200,205}

\def\wacvPaperID{1116} 
\def\confName{WACV}
\def\confYear{2026}

\title{ChartQA-X: Generating Explanations for Visual Chart Reasoning}

\author{Shamanthak Hegde \hspace{10pt} Pooyan Fazli \hspace{10pt} Hasti Seifi \\
  Arizona State University \\
  {\small\tt \{shamanthak,pooyan,hasti.seifi\}@asu.edu} \\
  {\normalsize \textcolor{cvprblue}{\url{https://teal-lab.github.io/chartqa-x}}}
  }

\begin{document}
\maketitle
\begin{abstract}
The ability to explain complex information from chart images is vital for effective data-driven decision-making. In this work, we address the challenge of generating detailed explanations alongside answering questions about charts. We present ChartQA-X, a comprehensive dataset comprising 30,799 chart samples across four chart types, each paired with contextually relevant questions, answers, and explanations. Explanations are generated and selected based on metrics such as faithfulness, informativeness, coherence, and perplexity. Our human evaluation with 245 participants shows that model-generated explanations in ChartQA-X surpass human-written explanations in accuracy and logic and are comparable in terms of clarity and overall quality.\ Moreover, models fine-tuned on ChartQA-X show substantial improvements across various metrics, including absolute gains of up to 24.57 points in explanation quality, 18.96 percentage points in question-answering accuracy, and 14.75 percentage points on unseen benchmarks for the same task. By integrating explanatory narratives with answers, our approach enables agents to convey complex visual information more effectively, improving comprehension and greater trust in the generated responses.

\end{abstract}

\section{Introduction}
\label{sec:intro}

\begin{figure}
    \centering
    \includegraphics[width=1.\linewidth]{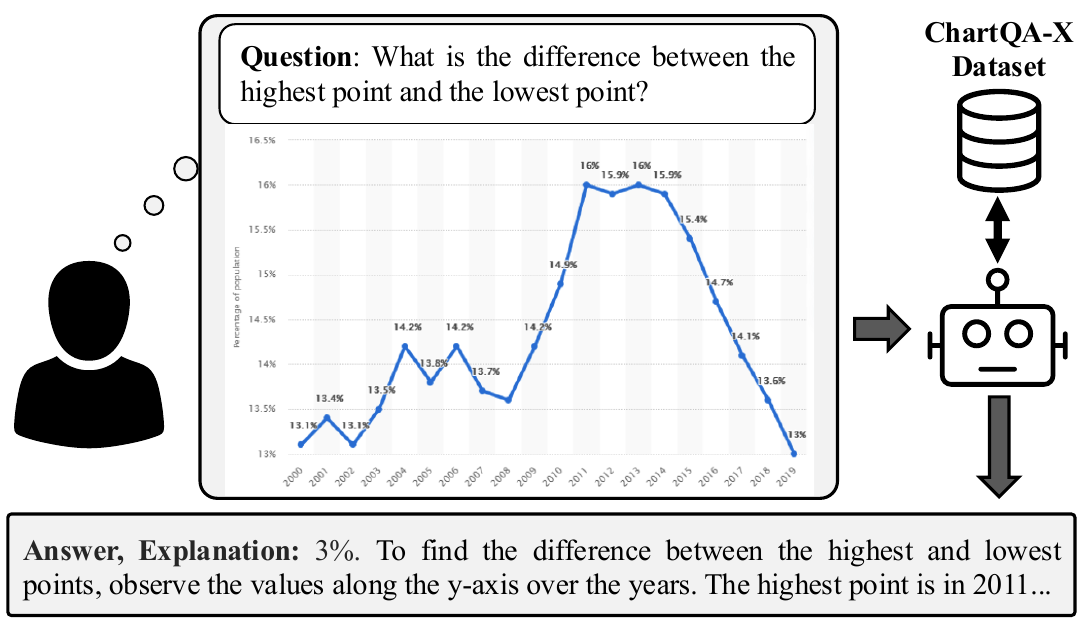}
    \caption{ChartQA-X dataset enables training VLMs capable of generating both answers and explanations in response to user questions about charts.}
    \label{fig:intro}
\end{figure}

Vision language models (VLMs) have made significant progress in addressing complex reasoning tasks~\cite{deepseek_r1, llavacot, llavanext, qwen2vl, molmo, internvl2}.
As their capabilities continue to grow, ensuring alignment with human values becomes increasingly critical.\ However, these models are often viewed as black boxes, producing answers without revealing the reasoning or evidence behind them. Providing explanations helps make their decision-making processes more transparent and interpretable to users.
Prior work on visual reasoning for images~\cite{beyondvqa, vcr} emphasizes the value of generating answers with explicit rationales 
to support model predictions. 
Others~\cite{vicor} leverage large language models for commonsense reasoning, demonstrating improved performance in visual question answering. 
While recent work has highlighted the value of explanations in vision-language reasoning tasks involving general images~\cite{exvqa,fsvqax, vqax, textvqax}, little research has explored explanation generation for data visualizations.\ Chart question answering (ChartQA) is an emerging area focused on developing models that can interpret chart data and answer user queries~\cite{chartqa,dvqa,figureqa,plotqa}.\ However, existing ChartQA models produce answers without the explanatory context needed to support user understanding and build trust. This limitation reduces their practical utility, as users often seek not only accurate answers but also clear and interpretable reasoning behind them.

Prior research often treats question answering and explanation generation as separate tasks~\cite{exvqa, pfte, nle, cqax}. Some approaches use one model to generate answers and a different model to produce explanations. Others focus solely on generating explanations without grounding them in the reasoning process that leads to an answer~\cite{faithful, vqa-e, e-vil}. This separation can disrupt the chain-of-thought (CoT) reasoning and result in inconsistencies between answers and their explanations. For example, a model might produce a correct explanation but an incorrect answer, or vice versa. These misalignments underscore the need for models that generate contextually relevant explanations in tandem with answering questions about chart data.\ Inspired by prior work~\cite{nlxgpt, unifying_vl}, we eliminate the need for separate models for answer prediction and explanation generation by using a single, general-purpose vision-language model to handle both tasks.\ This unified approach ensures that explanations are closely aligned with the predicted answers while also reducing memory usage and inference time, as shown in~\cite{nlxgpt}.

Motivated by this goal, we introduce \textbf{ChartQA-X}, a comprehensive dataset comprising 30,799 chart samples (28,299 train and 2,500 test), each paired with contextually relevant questions, answers, and explanations. Explanations are selected from the outputs of six state-of-the-art vision-language models. Figure~\ref{fig:intro} illustrates the ChartQA-X task setup, demonstrating how the dataset enables a model to generate both the answer and an explanation for a given chart-based question. While recent VLMs can generate explanations, their quality varies widely across questions and chart types. Without additional supervision, outputs are often verbose, ungrounded, or inconsistent. To ensure high-quality explanations in ChartQA-X, we simplify the task during dataset creation by supplying full context (e.g., answer, data table) and generating diverse candidate explanations with multiple VLMs. These are then filtered using automatic reasoning metrics to retain only high-quality explanations.  
ChartQA-X includes a diverse range of chart types, including (1) horizontal bar charts, (2) vertical bar charts, (3) line charts, and (4) pie charts, all derived from the original ChartQA dataset~\cite{chartqa}. 
We assess ChartQA-X through an online human-subject study with 245 participants who rate ChartQA-X explanations as comparable to or better than human-written ones in terms of accuracy, clarity, logic, and overall quality. 

By releasing a validated dataset, ChartQA-X enables benchmarking and facilitates progress in VLM explanation generation for chart questions. 
To establish performance baselines, we fine-tune three state-of-the-art vision-language models, i.e., LLaVA 1.6~\cite{llavanext}, Qwen2-VL~\cite{qwen2vl}, and InternVL-2.5~\cite{internvl25}, on the ChartQA-X dataset and evaluate them in terms of the quality of generating explanations and accuracy of question answering. We further assess their question-answering capabilities on unseen benchmark chart datasets, including DVQA~\cite{dvqa}, PlotQA~\cite{plotqa}, and FigureQA~\cite{figureqa}.
In summary, our contributions are as follows:

\begin{itemize}
    \item We introduce ChartQA-X, the largest dataset containing 30,799 chart samples with detailed explanations spanning diverse chart types.
    \item We provide empirical validation of the dataset through an online study involving 245 human participants, showing that ChartQA-X explanations are comparable to or surpass human-written explanations across four metrics. 
    \item Models fine-tuned on ChartQA-X demonstrate significant improvements across various metrics, including an absolute gain of up to 24.57 points in explanation quality, 18.96 percentage points in question-answering accuracy, and 14.75 percentage points on unseen benchmark datasets for the same task.
\end{itemize}

\section{Related Work}
\label{sec:related_works}


\begin{figure*}[t!]
    \centering
    \includegraphics[width=1.\linewidth]{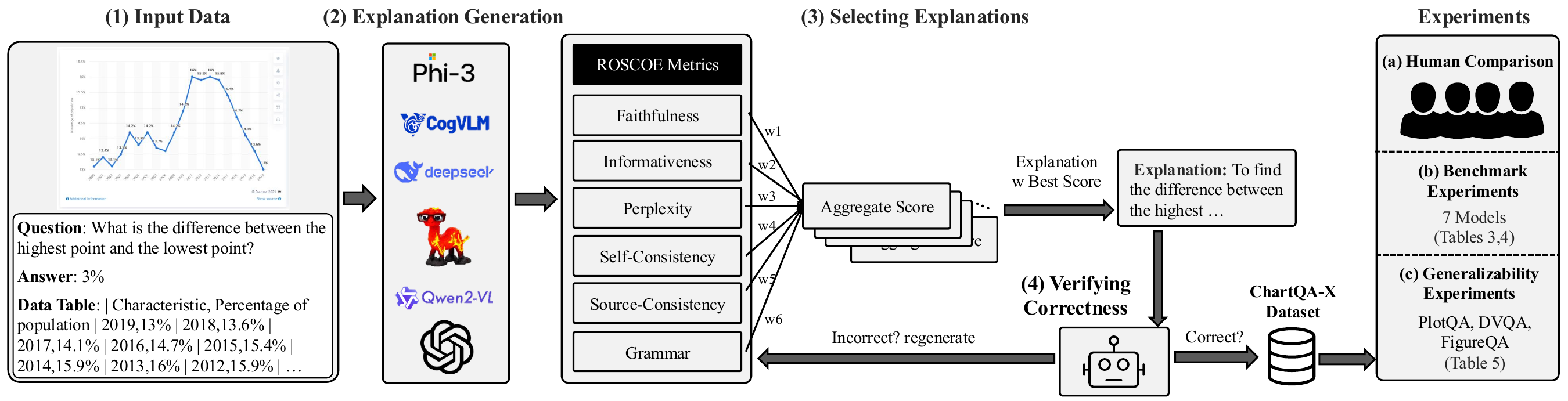}
    \caption{ChartQA-X is constructed in four stages: (1) preparing input data, (2) generating explanations using six VLMs, (3) selecting high-quality explanations based on ROSCOE scores, and (4) verifying explanation correctness. The dataset is evaluated through three types of experiments: (a) human-subject studies, (b) benchmark evaluations using accuracy and text generation metrics, and (c) generalizability tests on unseen datasets.}

    \label{fig:prompt_eng}
\end{figure*}

\subsection{Explainable AI for Visual Data}

Early work on natural language explanations focuses on vision tasks such as image classification~\cite{nle}. This line of research is later extended to language tasks~\cite{e-snli, pfte, wt5, explain} and vision-language tasks, including visual question answering (VQA)~\cite{e-vil, vqa-e, vqax, faithful}. For example, \citet{vqax} introduce two datasets: VQA-X and ACT-X. VQA-X augments the VQAv2 dataset~\cite{VQAv2} with natural language explanations that justify the answers provided by VQA models, while ACT-X provides explanations for activity recognition tasks. \citet{nlrwfsvr} use separate models to extract visual features from images, including object bounding boxes, role boxes with coordinates, and VisualCOMET embeddings~\cite{jsl}. These features, along with the question and ground truth answer, are fed into GPT-2~\cite{GPT-2} to generate an explanation. However, this approach treats question answering and explanation generation as separate processes. NLX-GPT~\cite{nlxgpt} addresses this limitation by using ResNet-101~\cite{resnet} and ViT~\cite{VIT} to extract image features, which are then combined with the question and answer into a single input sequence. This sequence is passed to a distilled version of the GPT-2 to generate both the answer and explanation. Similarly, we also train a single model to perform both tasks in a unified manner, but we focus on chart data.

\subsection{Chart-Based Question Answering}

Chart-based question answering has received increasing attention, leading to the development of several benchmark datasets, including FigureQA~\cite{figureqa}, DVQA~\cite{dvqa}, LEAF-QA~\cite{leafqa}, and LEAF-QA++\cite{leafqa++}. FigureQA and DVQA feature charts generated from synthetic data, whereas LEAF-QA and LEAF-QA++ are based on real-world data. These datasets typically rely on a small set of question templates, and answers are limited to a fixed vocabulary. In contrast, PlotQA\cite{plotqa} introduces open-vocabulary questions that often require performing mathematical operations on chart data.  ChartQA~\cite{chartqa} expands the scope of chart QA by providing a large-scale dataset with 9,608 human-written questions and 23,111 automatically generated ones derived from chart summaries using the T5 model~\cite{t5}. In terms of explanation, \citet{cqax} present a small, manually constructed dataset comprising 52 charts and 629 question–answer pairs to examine how humans reason and explain when answering chart-related questions. This remains one of the few publicly available explanation datasets in this domain. Building on this prior work, we introduce the largest explanation dataset to date for chart question answering.

\section{ChartQA-X}
\label{methodology}

Figure~\ref{fig:prompt_eng} illustrates the process of constructing the ChartQA-X dataset, which involves four main steps: (1) preparing the input data, (2) generating explanations using six VLMs, (3) evaluating and selecting the best explanation for each question-answer based on various metrics, and (4) verifying the correctness of the selected explanations. We describe each of these steps in detail below.

\subsection{Input Data} 
\label{sec:input}
\begin{figure}[t]
    \centering
    \includegraphics[width=1.0\linewidth]{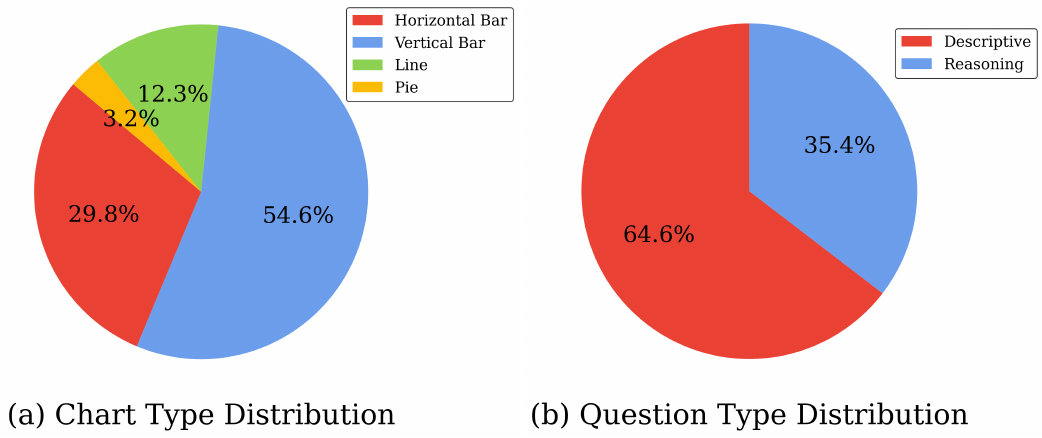}
    \caption{Distribution of (a) chart types, and (b) question types in the dataset.}
    \label{fig:dist_data}
\end{figure}

ChartQA-X is built upon the ChartQA dataset~\cite{chartqa}, which consists of 18,317 chart images and 30,799 question–answer pairs. It spans four chart types and features a diverse range of descriptive and reasoning questions (Figure~\ref{fig:dist_data}) with varying levels of complexity.\ Reasoning questions require calculation using chart data, while descriptive questions involve directly reading values from the chart. Each data point in the ChartQA-X dataset consists of four components: 
(1) Chart Image ($I$): A visual representation of the data, which may take the form of horizontal bar charts, vertical bar charts, line charts, or pie charts. The image provides visual context for the model, enabling it to interpret the data and incorporate visual attributes, such as color and shape, into its explanations.
(2) Data Table ($D$): A structured, tabular representation of the chart’s underlying data, offering precise numerical values and categorical labels. This component complements the visual chart by providing the chart data in a machine-readable format.
(3) Question ($Q$): A natural language query about the information presented in the chart. Questions span a variety of types, including trend analysis, comparisons, arithmetic reasoning, and specific data retrieval, serving as the primary prompt for the model.
(4) Answer ($A$): The correct response to the question, derived from the chart and its associated table. This is the expected output the model must produce, and the explanation is evaluated based on how well it supports this answer.


\subsection{Explanation Generation}
We generate explanations for each question-answer pair using six different VLMs: LLaVA 1.6~\cite{llavanext}, Phi-3~\cite{phi3}, CogVLM~\cite{cogvlm}, Deepseek-VL~\cite{deepseekvl}, Qwen2-VL~\cite{qwen2vl}, and GPT-4o~\cite{gpt-4o}. Each model is prompted to generate explanations in a chain-of-thought reasoning format~\cite{cot}, which breaks down the reasoning process into a sequence of logical steps, making the model's thought process transparent.\ This involves analyzing the chart to identify key data points, trends, and relationships, aligning the relevant information with the question, and constructing a coherent narrative that links the data to the final answer. Our final prompt is based on iterative experimentation with several prompt designs to elicit high-quality explanations from VLMs. Our early prompt versions include a fixed set of sub-questions to guide step-wise reasoning, such as \emph{``What values are shown?''} or \emph{``What comparisons are relevant?''}. While structured, this approach lacks flexibility and performs poorly across diverse question types. We next attempt an open-ended prompt that instructs the model to generate its own sub-questions before answering them. This produces longer outputs, but often includes irrelevant or redundant steps and conversational phrasing. To improve clarity and relevance, our final prompt is designed following HCI literature to emphasize concise reasoning with visual references like human explanations~\cite{cqax}, limit responses to four sentences, and avoid conversational language. Specifically, 
we design the following prompt:\\

\vspace{0.2cm}
\setlength{\fboxsep}{8pt} 
\setlength{\fboxrule}{0.3mm}
\noindent
\fcolorbox{black}{gray!15}{ 
    \begin{minipage}{0.88\linewidth}
    \raggedright
 \justify{\textit{Think like a human to arrive at the given answer. Generate an explanation (no more than four sentences) that outlines the steps taken to derive the answer using chain-of-thought reasoning. Focus on visual elements in the chart, such as color, height, position, and labels, and avoid conversational language or unnecessary commentary.}}
    \end{minipage}
}
\vspace{0.5cm}

\noindent 
We note that explanation generation during dataset construction is not equivalent to explanation generation at inference time. During this phase, we provide models with full context (the chart, question, answer, and data table), which significantly lowers task complexity. Furthermore, to reduce model-specific bias, we sample explanations from six different VLMs. 
As a result, the final dataset is composed of explanations not tied to a single model's reasoning style. This strategy mitigates circularity and enables models to learn from superior or complementary reasoning patterns present across models.





\subsection{Evaluating and Selecting Explanations}

\begin{table}[b]
\centering
\renewcommand{\arraystretch}{1.5}
\begin{adjustbox}{width=\columnwidth,center}
\begin{tabular}{@{}@{\hskip 0.2cm}ll@{\hskip 0.2cm}@{}}
\toprule
\textbf{Metric} & \textbf{Equation} \\
\midrule
FS  & \( \frac{1}{N} \sum_{i=1}^{N} r\text{-align}(h_i \rightarrow s) \) \\
FT  & \( \frac{1}{N + M} \sum_{i=1}^{N} \left[ r\text{-align}(h_i \rightarrow s) \right] + \sum_{j=1}^{M_i} r\text{-align}^{\text{token}}(h_{i,j} \rightarrow s) \) \\
IS  & \( \left( \frac{1}{T} \sum_{t=1}^{T} r\text{-align}(s_t \rightarrow h) + \frac{1}{N} \sum_{i=1}^{N} r\text{-align}(h_i \rightarrow s) \right) \bigg/ 2 \) \\
IC  & \( \left( 1 + \cos(h, s) \right) \big/ 2 \) \\
SRC & \( 1 - \max_{i = 1 \ldots N} \max_{j = 1 \ldots T} p_{\text{contradicts}}(h_i, s_j) \) \\
SFC & \( 1 - \max_{i = 2 \ldots N} \max_{j < i} p_{\text{contradicts}}(h_i, h_j) \) \\
PS  & \( \frac{1}{\frac{1}{N} \sum_{i=1}^{N} \text{PPL}(h_i)} \) \\
PC  & \( \frac{1}{\text{PPL}(h)} \) \\
GS  & \( \frac{1}{N} \sum_{i=1}^{N} p_{\text{gram}}(h_i) \) \\
\bottomrule
\end{tabular}
\end{adjustbox}
\caption{Equations for calculating ROSCOE metrics~\cite{roscoe}.}
\label{tab:roscoe_eq}
\end{table}

\begin{table*}[t]
\centering
\scriptsize
\begin{adjustbox}{width=1.0\textwidth,center}
\begin{tabular}{lccccccccc|c}
\toprule
\textbf{Models}    & \textbf{FS}     & \textbf{FT}     & \textbf{IS}     & \textbf{IC}     & \textbf{SRC}    & \textbf{SFC}    & \textbf{PS}     & \textbf{PC}     & \textbf{GS}  & \textbf{AS}   \\ 
    \midrule
Phi-3~\cite{phi3}       & 0.845          & 0.933          & 0.860          & 0.943          & 0.673          & 0.665          &    \textbf{0.993}    &   0.916          & 0.943   &  0.856       \\
CogVLM~\cite{cogvlm}     & 0.850          & 0.929          & 0.852          & 0.937          & 0.607          & 0.587          & 0.988          & 0.937          & 0.937     & 0.840     \\
Deepseek-VL~\cite{deepseekvl} & 0.845          & 0.933          & 0.860          & 0.943          & 0.673          & 0.665          & \textbf{0.993} & 0.916          & 0.943    & 0.856      \\
LLaVA 1.6~\cite{llavanext}   & \textbf{0.863}          & 0.935          & 0.860          & 0.956          & 0.737          & 0.552          & 0.989          & 0.882          & 0.956    & 0.855      \\
Qwen2-VL~\cite{qwen2vl}     & \textbf{0.863} & 0.936          & 0.867          & 0.952          & 0.737       & 0.655          & 0.984          & 0.897          & \textbf{0.960}  &  0.867  \\
GPT-4o~\cite{gpt-4o}      & 0.840          & \textbf{0.937} & \textbf{0.883} & \textbf{0.958} & \textbf{0.791} & \textbf{0.773} & \textbf{0.993}          & \textbf{0.948} & 0.921   & \textbf{0.884}       \\ 
\bottomrule
\end{tabular}
\end{adjustbox}
\caption{ROSCOE scores for six VLMs on the ChartQA-X dataset. Faithfulness Step (FS), Faithfulness Token (FT), Informativeness Step (IS), Informativeness Chain (IC), Source-Consistency (SRC), Self-Consistency (SFC), Perplexity Step (PS),  Perplexity Chain (PC), Grammar Step (GS), and Aggregate Score (AS).}
\label{tab:gen_scores}
\end{table*}

To assess the quality and effectiveness of the explanations, we use the ROSCOE evaluation suite~\cite{roscoe}, which is specifically designed to score step-by-step reasoning. ROSCOE offers a comprehensive framework that evaluates reasoning quality across nine critical dimensions. In our setup, we define the \textit{source} context as $s=\{s_1,..., s_T \}$, a sequence of $T$ sentences consisting of the instruction followed by the question. The \textit{hypothesis} is denoted as $h=\{h_1,...,h_N\}$, a sequence of $N$ reasoning steps, including the final explanation. Reasoning Alignment ($r{-align}$) measures the degree to which each step in a multi-step explanation aligns with the source context (i.e., question). Alignment for each step is computed as: 
\begin{equation}
r\text{-align}(h_i \to s) = \frac{[1 + \max_{j=1}^T(\cos(h_i, s_j)]}{2}
\label{eq: ralign}
\end{equation}
yielding a vector $\in [0, 1]^N$. This reflects the semantic closeness of each reasoning step to the source. Contradiction Probability ($p_\text{contradicts}$) is estimated using a Natural Language Inference (NLI) model, in this case a fine-tuned DeBERTa-v3~\cite{debertav3_nli}, that classifies explanation-source pairs into entailment, neutral, or contradiction. The $p_\text{contradicts}$ score is the model's confidence that the explanation contradicts the source, serving as a proxy for logical inconsistency. English grammatical acceptability $p_{gram}$ is scored by a classifier model~\cite{gram_class}. Table~\ref{tab:roscoe_eq} provides the equations for each of the nine ROSCOE metrics.

\vspace{0.08cm}
\noindent \textbf{1. Faithfulness Step (FS)} evaluates whether the model misinterprets the problem statement or if the reasoning chain is vague, irrelevant, or misuses information. It measures the alignment ($r{-align}$) between each step of the hypothesis ($h_i$) and source context ($s$) and is computed as the mean alignment score across all steps.

\vspace{0.08cm}
\noindent \textbf{2. Faithfulness Token (FT)} measures token-level similarity based on embedding alignment. Given $M_i$ tokens in step $h_i$, where $h_{i,j}$ is the $j$th token in the $i$th step, $r{-align}^{token}$ denotes the alignment vector from tokens in $h_i$   to all tokens in the source context $s$.

\vspace{0.08cm}
\noindent \textbf{3. Informativeness Step (IS)} measures how effectively each reasoning step uses information from the source context.

\vspace{0.08cm}
\noindent \textbf{4. Informativeness Chain (IC)} quantifies how well the overall reasoning chain ($h$) aligns with the source context ($s$).

\vspace{0.08cm}
\noindent \textbf{5. Source-Consistency (SRC)} measures logical entailment errors between the generated reasoning $h$ and the source context $s$.

\vspace{0.08cm}
\noindent \textbf{6. Self-Consistency (SFC)} measures logical entailment errors within a reasoning step, capturing its internal coherence.

\vspace{0.08cm}
\noindent \textbf{7. Perplexity Step (PS)} is the average perplexity (PPL) of all tokens in the generated reasoning steps, where each token is scored using only the preceding tokens within the same step. Language coherence is evaluated using the GPT2-Large model~\cite{GPT-2} to compute PPL.

\vspace{0.08cm}
\noindent \textbf{8. Perplexity Chain (PC)} is the average perplexity (PPL) of all tokens in the generated reasoning steps. 

\vspace{0.08cm}
\noindent \textbf{9. Grammar Step (GS)} evaluates the grammatical correctness of each individual step in an explanation, focusing on sentence structure, verb tense, punctuation, and other grammatical aspects within each reasoning unit.

\begin{figure*}[t]
    \centering
    \includegraphics[width=1.0\linewidth]{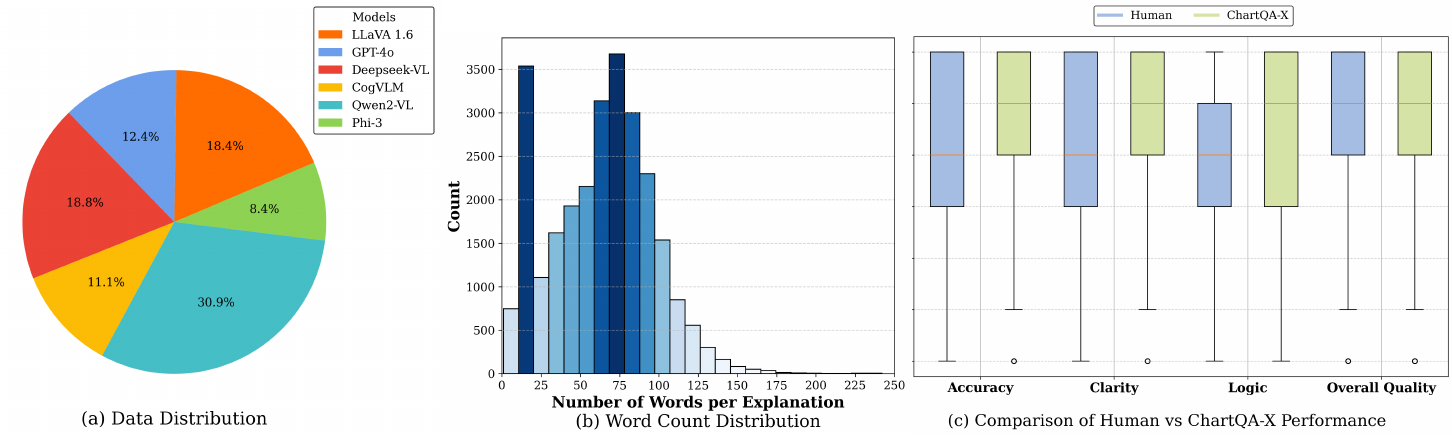}
    \caption{Distribution of explanations in ChartQA-X including (a) percentage of explanations obtained from each VLM, (b) lengths of explanations, and (c) Box plot comparing Human and ChartQA-X performance across four evaluation metrics: Accuracy, Clarity, Logic, and Overall Quality. Each pair of boxes represents the distribution of scores for a specific metric, highlighting differences in performance and variance.}
    \label{fig:dist}
\end{figure*}

Each explanation generated by the six VLMs is evaluated using the nine ROSCOE metrics. Since different models perform best on different subsets of the dataset, we select, for each sample, the explanation with the highest overall score across all metrics. To ensure fair comparison, all metric scores are normalized to a $0$-$1$ scale, with 1 indicating the best performance and 0 indicating the worst. For perplexity-based metrics (PS and PC), where lower is better, scores are inverted by subtracting them from 1. 
We then apply weights between 0 and 1 to each metric to compute an aggregate score. 
To prioritize alignment between the explanation, question, and answer, we assign higher weights to faithfulness and informativeness metrics and lower weights to others. We set the weights as follows: 0.2 for the faithfulness step, 0.15 each for the faithfulness token and the informativeness chain, 0.1 for self-consistency and source-consistency, and 0.05 each for the perplexity step, perplexity chain, and grammar step. This weighting scheme emphasizes explanation quality and relevance while also considering coherence, fluency, and grammatical accuracy. 
 To evaluate the stability of this approach, we conduct a sensitivity analysis using a variety of alternative weight configurations, including uniform, random, and metric-focused schemes. The results, presented in Appendix~\ref{weight_sens}, show that while weighting affects selection, our choices align well with other meaningful configurations. 

We compute an Aggregate Score (AS) for each explanation as a weighted sum of all normalized metric scores.
For each sample, we select the explanation with the highest \textit{Aggregate Score (AS)} among the six models, resulting in a high-quality collection of explanations. Table~\ref{tab:gen_scores} presents the average ROSCOE scores across all ChartQA samples for each of the six VLMs.

\begin{table*}[!t]
\scriptsize
\begin{adjustbox}{width=1.0\textwidth,center}
\begin{tabular}{lcccc|cc|c}
\toprule
\multirow{3}{*}{\textbf{Models}} 
& \multicolumn{4}{c}{\textbf{Chart Types}} 
& \multicolumn{2}{c}{\textbf{Question Types}} 
& \multirow{3}{*}{\textbf{Overall}} \\
\cmidrule(lr){2-5} \cmidrule(lr){6-7}
& \textbf{Horizontal Bar} & \textbf{Vertical Bar} & \textbf{Pie} & \textbf{Line} 
& \textbf{Descriptive} & \textbf{Reasoning} \\
\midrule
\rowcolor{gray!15}\multicolumn{8}{c}{\textit{State-of-the-Art Models}}\\
GPT-4o~\cite{gpt-4o}                           & 75.03 & 71.23 & \underline{82.56} & 66.33 & 70.15 & \underline{75.11} & 73.40 \\
Deepseek-VL~\cite{deepseekvl}                      & 52.01 & 50.09 & 46.51 & 45.14 & 55.52 & 42.54 & 48.64 \\
CogVLM~\cite{cogvlm}                           & 56.53 & 54.59 & 54.07 & 48.63 & 56.69 & 51.19 & 53.62 \\
Phi-3~\cite{phi3}                            & 74.51 & \underline{79.90} & 77.33 & \underline{69.58} & \underline{81.13} & 70.70 & \underline{75.53} \\

\midrule
\rowcolor{gray!15}\multicolumn{8}{c}{\textit{Models Fine-tuned on ChartQA-X}}\\
LLaVA-1.6~\cite{llavanext}                        & 23.93 & 17.50 & 34.30 & 23.44 & 22.46 & 20.56 & 23.53 \\
\rowcolor{yellow!10}\hspace{2mm} + \textbf{ChartQA-X} & 35.32 $\scriptstyle\textcolor{darkgreen}{+11.39}$ & 29.55 $\scriptstyle\textcolor{darkgreen}{+12.05}$ & 47.67 $\scriptstyle\textcolor{darkgreen}{+13.37}$ & 28.43 $\scriptstyle\textcolor{darkgreen}{+4.99}$ & 34.75 $\scriptstyle\textcolor{darkgreen}{+12.29}$ & 29.57 $\scriptstyle\textcolor{darkgreen}{+9.01}$ & 34.22 $\scriptstyle\textcolor{darkgreen}{+10.69}$ \\
Qwen2-VL~\cite{qwen2vl}                         & 42.82 & 36.57 & 50.58 & 34.16 & 43.01 & 34.33 & 40.25 \\
\rowcolor{yellow!10}\hspace{2mm} + \textbf{ChartQA-X} & 53.56 $\scriptstyle\textcolor{darkgreen}{+10.74}$ & 55.46 $\scriptstyle\textcolor{darkgreen}{+18.89}$ & 54.07 $\scriptstyle\textcolor{darkgreen}{+3.49}$ & 53.12 $\scriptstyle\textcolor{darkgreen}{+18.96}$ & 59.69 $\scriptstyle\textcolor{darkgreen}{+16.68}$ & 48.01 $\scriptstyle\textcolor{darkgreen}{+13.68}$ & 53.99 $\scriptstyle\textcolor{darkgreen}{+13.74}$\\
InternVL-2.5~\cite{internvl2}                     & \underline{75.81} & 76.43 & 73.26 & 68.83 & 80.25 & 68.23 & 73.80 \\ 
\rowcolor{yellow!10}\hspace{2mm} + \textbf{ChartQA-X} & \textbf{80.98} $\scriptstyle\textcolor{darkgreen}{+5.17}$ & \textbf{82.76} $\scriptstyle\textcolor{darkgreen}{+6.33}$ & \textbf{84.30} $\scriptstyle\textcolor{darkgreen}{+11.04}$ & \textbf{70.57} $\scriptstyle\textcolor{darkgreen}{+1.74}$ & \textbf{83.69} $\scriptstyle\textcolor{darkgreen}{+3.44}$ & \textbf{76.35} $\scriptstyle\textcolor{darkgreen}{+8.12}$ & \textbf{79.78} $\scriptstyle\textcolor{darkgreen}{+5.98}$ \\
\bottomrule
\end{tabular}
\end{adjustbox}
\caption{Question answering accuracy (\%), without data table in the input, calculated for different chart and question types (test set only). Best scores are in \textbf{bold}, and second-best scores are \underline{underlined}.}
\label{tab:cat_table}
\end{table*}

\begin{table*}[!t]
\scriptsize
\begin{adjustbox}{width=1.0\textwidth,center}
\begin{tabular}{lcccc|cc|c}
\toprule
\multirow{3}{*}{\textbf{Models}} 
& \multicolumn{4}{c}{\textbf{Chart Types}} 
& \multicolumn{2}{c}{\textbf{Question Types}} 
& \multirow{3}{*}{\textbf{Overall}} \\
\cmidrule(lr){2-5} \cmidrule(lr){6-7}
& \textbf{Horizontal Bar} & \textbf{Vertical Bar} & \textbf{Pie} & \textbf{Line} 
& \textbf{Descriptive} & \textbf{Reasoning} \\
\midrule
\rowcolor{gray!15}\multicolumn{8}{c}{\textit{State-of-the-Art Models}}\\
GPT-4o~\cite{gpt-4o}                          & \underline{76.45} & 76.43 & \textbf{84.88} & 70.07 & 75.74 & \textbf{76.31} & \underline{76.64} \\
Deepseek-VL~\cite{deepseekvl}                  & 59.07 & 62.26 & 56.98 & 52.37 & 68.20 & 48.87 & 57.96 \\
CogVLM~\cite{cogvlm}                           & 60.74 & 61.74 & 65.12 & 50.62 & 63.68 & 55.40 & 59.55 \\
Phi-3~\cite{phi3}                            & 74.65 & \underline{82.96} & 74.42 & \underline{72.32} & 84.47 & 70.56 & 76.56 \\
\midrule
\rowcolor{gray!15}\multicolumn{8}{c}{\textit{Models Fine-tuned on ChartQA-X}}\\
LLaVA-1.6~\cite{llavanext}                        & 48.39 & 55.74 & 53.49 & 42.14 & 57.40 & 43.73 & 50.15 \\
\rowcolor{yellow!10}\hspace{2mm} + \textbf{ChartQA-X} & 60.36 $\scriptstyle\textcolor{darkgreen}{+11.97}$ & 62.70 $\scriptstyle\textcolor{darkgreen}{+6.96}$ & 63.95 $\scriptstyle\textcolor{darkgreen}{+10.46}$ & 52.87 $\scriptstyle\textcolor{darkgreen}{+10.73}$ & 67.83 $\scriptstyle\textcolor{darkgreen}{+10.43}$ & 51.83 $\scriptstyle\textcolor{darkgreen}{+8.10}$ & 59.92 $\scriptstyle\textcolor{darkgreen}{+9.77}$ \\
Qwen2-VL~\cite{qwen2vl}                         & 72.90 & 77.61 & 72.49 & 65.83 & 82.02 & 62.94 & 72.30 \\
\rowcolor{yellow!10}\hspace{2mm} + \textbf{ChartQA-X} & 74.90 $\scriptstyle\textcolor{darkgreen}{+2.0}$ & 80.95 $\scriptstyle\textcolor{darkgreen}{+3.34}$ & 77.91 $\scriptstyle\textcolor{darkgreen}{+5.42}$ & 67.83 $\scriptstyle\textcolor{darkgreen}{+2.0}$ & \underline{85.72} $\scriptstyle\textcolor{darkgreen}{+3.70}$ & 66.20 $\scriptstyle\textcolor{darkgreen}{+3.26}$ & 75.59 $\scriptstyle\textcolor{darkgreen}{+3.29}$\\
InternVL-2.5~\cite{internvl2}                     & 75.42 & 80.35 & 76.16 & 71.57 & 85.65 & 67.07 & 76.04 \\ 
\rowcolor{yellow!10}\hspace{2mm} + \textbf{ChartQA-X} & \textbf{79.28} $\scriptstyle\textcolor{darkgreen}{+3.86}$ & \textbf{83.74} $\scriptstyle\textcolor{darkgreen}{+3.39}$ & \underline{83.72} $\scriptstyle\textcolor{darkgreen}{+7.56}$ & \textbf{72.82} $\scriptstyle\textcolor{darkgreen}{+1.25}$ & \textbf{87.35} $\scriptstyle\textcolor{darkgreen}{+1.70}$ & \underline{72.65} $\scriptstyle\textcolor{darkgreen}{+5.58}$ & \textbf{79.93} $\scriptstyle\textcolor{darkgreen}{+3.89}$ \\
\bottomrule
\end{tabular}
\end{adjustbox}
\caption{Question answering accuracy (\%), with data table included in the input, calculated for different chart and question types (test set only). Best scores are in \textbf{bold}, and second-best scores are \underline{underlined}.}
\label{tab:cat_dt_table}
\end{table*}

\begin{table*}[ht]
 \begin{adjustbox}{width=\textwidth,center}
\begin{tabular}{lccccccccc|c}
\toprule
\textbf{Models}    & \textbf{FS}     & \textbf{FT}     & \textbf{IS}     & \textbf{IC}     & \textbf{SRC}    & \textbf{SFC}    & \textbf{PS}     & \textbf{PC}     & \textbf{GS}  & \textbf{AS}   \\ 
\midrule
\rowcolor{gray!15}\multicolumn{11}{c}{\textit{State-of-the-Art Models}}\\
Phi-3~\cite{phi3}             &  0.865  &  0.939  &  0.873  &  \underline{0.925}  &  0.678  &  0.792  &  0.995  &  0.947  &  0.946  &  0.875 \\
CogVLM~\cite{cogvlm}            &  0.873  &  0.938  &  0.887  &  0.919  &  0.471  &  0.714  &  \textbf{0.997}  &  \textbf{0.997}  &  0.937  &  0.851 \\
Deepseek-VL~\cite{deepseekvl}       &  0.871  &  0.939  &  0.884  &  0.921  &  0.311  &  0.722  &  \underline{0.996}  &  0.928  &  0.928  &  0.832 \\
GPT-4o~\cite{gpt-4o}            &  0.882  &  0.941  &  \underline{0.892}  &  0.916  &  \underline{0.803}  &  0.850  &  0.972  &  0.972  &  0.937  &  0.899 \\
\midrule
\rowcolor{gray!15}\multicolumn{11}{c}{\textit{Models Fine-tuned on ChartQA-X}}\\
InternVL-2.5~\cite{internvl2}      &  0.862  &  0.939  &  0.868  &  0.918  &  0.715  &  0.722  &  0.995  &  0.959  &  0.951  &  0.870 \\
\rowcolor{yellow!10}\hspace{2mm} + \textbf{ChartQA-X}  &  
0.894 $\scriptstyle\textcolor{darkgreen}{+0.032}$  &  
0.947 $\scriptstyle\textcolor{darkgreen}{+0.008}$  &  
0.877 $\scriptstyle\textcolor{darkgreen}{+0.009}$  &  
0.922 $\scriptstyle\textcolor{darkgreen}{+0.004}$  &  
0.797 $\scriptstyle\textcolor{darkgreen}{+0.082}$  &  
0.853 $\scriptstyle\textcolor{darkgreen}{+0.131}$  &  
0.994 $\scriptstyle\textcolor{darkgreen}{-0.001}$  &  
0.963 $\scriptstyle\textcolor{darkgreen}{+0.004}$  &  
0.958 $\scriptstyle\textcolor{darkgreen}{+0.007}$  &  
\underline{0.901} $\scriptstyle\textcolor{darkgreen}{+0.031}$ \\
LLaVA 1.6~\cite{llavanext}         &  0.884  &  0.945  &  0.877  &  0.901  &  0.515  &  0.858  &  0.989  &  0.991  &  0.941  &  0.869 \\
\rowcolor{yellow!10}\hspace{2mm} + \textbf{ChartQA-X}  &  
\textbf{0.928} $\scriptstyle\textcolor{darkgreen}{+0.044}$  &  
\underline{0.949} $\scriptstyle\textcolor{darkgreen}{+0.004}$  &  
\textbf{0.894} $\scriptstyle\textcolor{darkgreen}{+0.017}$  &  
0.918 $\scriptstyle\textcolor{darkgreen}{+0.017}$  &  
0.609 $\scriptstyle\textcolor{darkgreen}{+0.094}$  &  
\textbf{0.920} $\scriptstyle\textcolor{darkgreen}{+0.062}$  &  
\textbf{0.997} $\scriptstyle\textcolor{darkgreen}{+0.008}$  &  
\underline{0.996} $\scriptstyle\textcolor{darkgreen}{+0.005}$  &  
\textbf{0.980} $\scriptstyle\textcolor{darkgreen}{+0.039}$  &  
\underline{0.901} $\scriptstyle\textcolor{darkgreen}{+0.032}$ \\
Qwen2-VL~\cite{qwen2vl}      &  0.868  &  0.941  &  0.868  &  0.918  &  0.674  &  0.871  &  0.983  &  0.995  &  0.964  &  0.884 \\
\rowcolor{yellow!10}\hspace{2mm} + \textbf{ChartQA-X}  &  
\underline{0.907} $\scriptstyle\textcolor{darkgreen}{+0.039}$  &  
\textbf{0.953} $\scriptstyle\textcolor{darkgreen}{+0.012}$  &  
0.884 $\scriptstyle\textcolor{darkgreen}{+0.016}$  &  
\textbf{0.963} $\scriptstyle\textcolor{darkgreen}{+0.045}$  &  
\textbf{0.902} $\scriptstyle\textcolor{darkgreen}{+0.228}$  &  
\underline{0.918} $\scriptstyle\textcolor{darkgreen}{+0.047}$  &  
\textbf{0.997} $\scriptstyle\textcolor{darkgreen}{+0.014}$  &  
\textbf{0.997} $\scriptstyle\textcolor{darkgreen}{+0.002}$  &  
\underline{0.970} $\scriptstyle\textcolor{darkgreen}{+0.006}$  &  
\textbf{0.932} $\scriptstyle\textcolor{darkgreen}{+0.048}$ \\
\bottomrule
\end{tabular}
\end{adjustbox}
\caption{ROSCOE scores on the ChartQA-X test set without data table in the input. Best scores are in \textbf{bold}, and second-best scores are \underline{underlined}. FS: Faithfulness Step, FT: Faithfulness Token, IS: Informativeness Step, IC: Informativeness Chain, SRC: Source-Consistency, SFC: Self-Consistency, PS: Perplexity Step,  PC: Perplexity Chain, GS: Grammar Step, and AS: Aggregate Score.}
\label{tab:bench_roscoe}
\end{table*}

\begin{table*}[!h]
\centering
\scriptsize
\begin{adjustbox}{width=0.75\textwidth,center}
\begin{tabular}{lcccc}
\toprule
\textbf{Models} & \textbf{BLEU-4} & \textbf{METEOR} & \textbf{ROUGE-L} & \textbf{CIDEr} \\
\midrule
\rowcolor{gray!15}\multicolumn{5}{c}{\textit{State-of-the-Art Models}}\\
GPT-4o~\cite{gpt-4o}      & 6.98             & 17.81            & 28.56            & 10.17           \\
Deepseek-VL~\cite{deepseekvl} & 1.18         & 6.58             & 8.02             & 7.16            \\
CogVLM~\cite{cogvlm}      & 1.03             & 9.91             & 19.22            & 3.58            \\
Phi-3~\cite{phi3}         & 12.83            & 20.76            & 30.06            & 30.18           \\
\midrule
\rowcolor{gray!15}\multicolumn{5}{c}{\textit{Models Fine-tuned on ChartQA-X}}\\
LLaVA 1.6~\cite{llavanext} & 11.07            & 16.40            & 26.44            & 29.19           \\
\rowcolor{yellow!10}\hspace{2mm} + \textbf{ChartQA-X}    & 17.95 $\scriptstyle\textcolor{darkgreen}{+6.88}$           & 23.60 $\scriptstyle\textcolor{darkgreen}{+7.2}$           & 33.52 $\scriptstyle\textcolor{darkgreen}{+7.08}$      & 33.83 $\scriptstyle\textcolor{darkgreen}{+4.64}$          \\
Qwen2-VL~\cite{qwen2vl}  & 6.49             & 15.18            & 25.81            & 12.31           \\
\rowcolor{yellow!10}\hspace{2mm} + \textbf{ChartQA-X}     & \textbf{20.50} $\scriptstyle\textcolor{darkgreen}{+14.01}$   & \textbf{24.31} $\scriptstyle\textcolor{darkgreen}{+9.13}$   & \textbf{34.69} $\scriptstyle\textcolor{darkgreen}{+8.88}$   & \underline{36.88} $\scriptstyle\textcolor{darkgreen}{+24.57}$          \\
InternVL-2.5~\cite{internvl25}  & 9.61      & 18.96            & 32.03            & 18.94           \\
\rowcolor{yellow!10}\hspace{2mm} + \textbf{ChartQA-X} & \underline{20.37} $\scriptstyle\textcolor{darkgreen}{+10.76}$ & \underline{23.71} $\scriptstyle\textcolor{darkgreen}{+4.75}$ & \underline{34.53} $\scriptstyle\textcolor{darkgreen}{+2.5}$ & \textbf{39.72} $\scriptstyle\textcolor{darkgreen}{+20.78}$  \\ 
\bottomrule
\end{tabular}
\end{adjustbox}

\caption{Evaluation results on n-gram metrics for the Chart-QA test set. Best scores are in \textbf{bold}, and second-best scores are \underline{underlined}.}
\label{tab:chartqax_result}
\end{table*}

\subsection{Verifying Correctness}
To verify the correctness of the top-scoring explanations, we leverage the same six state-of-the-art vision-language models: LLaVA 1.6, Phi-3, CogVLM, Deepseek-VL, Qwen2-VL, and GPT-4o. For each explanation generated by a given model, the remaining five models serve as independent verifiers. Each verifier receives the input chart, question, ground-truth answer, data table, and candidate explanation, and produces a binary judgment, Correct or Incorrect. An explanation is accepted into the ChartQA-X dataset if a majority (i.e., at least three out of five) of the verifying models label it as correct. If the explanation fails to reach majority approval, we move to the next highest-scoring explanation based on its ROSCOE score and repeat the same verification process. This iterative evaluation continues until an explanation is accepted. If none of the initial explanations are accepted, we regenerate the candidate explanations from all six models and restart the process. Figure~\ref{fig:dist}(a) shows the contribution of each model to the ChartQA-X dataset, while Figure~\ref{fig:dist}(b) presents the word count distribution of the accepted explanations. See Appendix~\ref{app:explen} for the distribution of explanation word counts across the six VLMs.

\section{Comparison to Human Explanations}
To assess the quality of ChartQA-X explanations in a realistic setting, we conducted a human-subject study involving 245 participants that compared ChartQA-X and human-written explanations.
We selected 180 samples from the ChartQA-X test set by sampling 30 instances from each of the six categories, defined by chart type (horizontal bar, vertical bar, pie, and line charts) and question type (descriptive and reasoning). This ensured a balanced coverage across formats and question types.

To collect human-written explanations, we recruited 10 graduate students outside the author team, each of whom was assigned 18 unique samples. These samples were carefully curated to include 3 examples from each of the 6 categories, ensuring comprehensive exposure to the dataset's diversity. The instructions closely mirrored the VLM’s input prompt: \textit{``Write an explanation (no more than
four sentences) that outlines the steps taken
to derive the answer. Focus on visual elements in the
chart, such as color, height, position, and labels, and avoid conversational language or
unnecessary commentary.''}\ Each human-written explanation was then paired with the corresponding ChartQA-X explanation, resulting in 180 paired comparison instances.


Each explanation pair was evaluated on Amazon Mechanical Turk by two different workers, with a total of 245 unique workers completing 360 tasks. Each task included a chart image, a data table, a question, an answer, and two unlabeled explanations (human, ChartQA-X), shown in random order. 
Workers rated each explanation using 1–7 Likert scales (1: poor, 4: neutral, 7: excellent) on four criteria: (a) Accurate: The explanation correctly reflects the chart data and supports the given answer, (b) Clear: The explanation is easy to understand and free from ambiguity, (c) Logical: The reasoning follows a coherent and sensible progression, and (d) Overall Quality: A holistic score accounting for usefulness, clarity, accuracy, and coherence. 

Figure~\ref{fig:dist}(c) shows the overall results of our user study. As seen in the figure, ChartQA-X explanations perform on par with human-written ones and slightly outperform them on average across all four criteria. In particular, they score significantly higher in accuracy ($p = 0.035$) and logical coherence ($p = 0.043$). These results suggest that ChartQA-X can generate high-quality explanations that closely match and sometimes surpass human reasoning. Example explanations and user ratings are provided in Appendix~\ref{app:examples}.

\section{Experiments and Evaluation}

\subsection{Baselines and Metrics}
We evaluate six open-source models, including LLaVA 1.6~\cite{llavanext}, Qwen2-VL~\cite{qwen2vl}, InternVL-2.5~\cite{internvl25}, Deepseek-VL~\cite{deepseekvl}, Phi-3~\cite{phi3}, and CogVLM~\cite{cogvlm}, and a proprietary model, GPT-4o~\cite{gpt-4o}, on the ChartQA-X test set.\ We also fine-tune LLaVA 1.6, Qwen2-VL, and InternVL-2.5 on the ChartQA-X training set to jointly generate both the answer and an accompanying explanation in a single output sequence, establishing three baseline models. Each training instance is formatted as: \textit{``Answer: \textless answer\textgreater. Explanation: \textless explanation\textgreater''}. The model is trained using a standard causal language modeling (LM) loss applied uniformly across the entire output sequence, without assigning distinct weights to the answer and explanation components. This joint generation approach promotes logical consistency between the answer and its explanation, enhancing interpretability, and reducing the risk of error propagation commonly associated with two-stage (answer-then-explanation) methods. At inference time, the model is prompted with:

\vspace{0.3cm}
\setlength{\fboxsep}{8pt} 
\setlength{\fboxrule}{0.3mm}
\noindent
\fcolorbox{black}{gray!15}{ 
    \begin{minipage}{0.88\linewidth}
    \raggedright
 \justify{\textit{Please answer the following question using the format below: The answer is: \textless answer\textgreater. Explanation: \textless concise, logical explanation justifying the answer\textgreater.}}
    \end{minipage}
}
\vspace{0.4cm}

\noindent This prompt is followed by the chart image and the question.
All experiments are conducted with a batch size of 1 on 4 NVIDIA A100 (80GB) GPUs. Fine-tuning is performed on 8 NVIDIA H100 (80GB) GPUs and takes approximately six hours. Inference time on the test set is about one hour for all models, except CogVLM, which requires roughly 2.5 hours. For consistency, we use 7B-parameter versions of all models, except InternVL, which has 8B.
We assess model performance using three criteria: (1) accuracy on the question-answering task, (2) explanation quality based on ROSCOE and n-gram metrics, including BLEU-4~\cite{bleu}, METEOR~\cite{meteor}, ROUGE-L~\cite{rouge-l}, and CIDEr~\cite{cider}, and (3) generalizability to unseen benchmarks after fine-tuning on ChartQA-X.

\subsection{Results}


Table~\ref{tab:cat_table} demonstrates the impact of fine-tuning on ChartQA-X across various chart and question types without data table in input. Fine-tuned models significantly outperform their base counterparts. InternVL-2.5 fine-tuned on ChartQA-X achieves the highest overall accuracy of 79.78\%. Substantial gains are observed across all chart types, with improvements of up to 18.96 percentage points (Qwen2-VL on line charts). On the reasoning subset, which is typically more challenging, ChartQA-X fine-tuning leads to notable boosts. For instance, Qwen2-VL improves from 34.33\% to 48.01\%. Even models with strong initial performance (e.g., InternVL-2.5) benefit, indicating the broad effectiveness of ChartQA-X. These results highlight ChartQA-X's value in enhancing both factual and inferential understanding of visual charts. While prior papers report higher accuracy on the original ChartQA benchmark, our evaluation shows lower performance due to a key difference in our setup: we do not provide the underlying data tables as input, which makes the task significantly more challenging. 
As shown in Table~\ref{tab:cat_dt_table}, incorporating the data table in the input substantially improves performance, with Qwen2-VL and LLaVA-1.6 achieving nearly twice their previous accuracy. 



Table~\ref{tab:bench_roscoe} further shows that fine-tuned models achieve significantly higher ROSCOE scores on the ChartQA-X test set. Table~\ref{tab:chartqax_result} shows that fine-tuning significantly improves explanation quality on the ChartQA-X test set, as measured by n-gram metrics. For example, Qwen2-VL shows significant absolute gains on CIDEr (+24.57), BLEU-4 (+14.01), and METEOR (+9.13), while InternVL-2.5 improves by +20.78, +10.76, and +4.75, respectively.

Finally, we evaluate the models' QA accuracy on a random subset of 20,000 samples from three other chart datasets, including DVQA~\cite{dvqa}, PlotQA~\cite{plotqa}, and FigureQA~\cite{figureqa}. As shown in Table~\ref{tab:ablat}, models fine-tuned on ChartQA-X show significant improvements up to +14.75\%.  Qwen2-VL performs best, reaching 97.32\% on DVQA and 95.47\% on FigureQA. 
Notably, FigureQA and PlotQA include dot-line plots, a chart type not present in ChartQA-X, highlighting the models’ ability to generalize to unseen chart types. ROSCOE score improvements for these datasets are reported in Appendix~\ref{app:results}.


\begin{table}[t]
\footnotesize
\begin{adjustbox}{width=\columnwidth,center}
\begin{tabular}{lccc}
\toprule
     \textbf{Models}    & \textbf{DVQA} & \textbf{PlotQA} & \textbf{FigureQA} \\ 
\midrule
\rowcolor{gray!15}\multicolumn{4}{c}{\textit{State-of-the-Art Models}}\\
Deepseek-VL~\cite{deepseekvl}        &   51.10      &    57.59     &   53.76    \\
CogVLM~\cite{cogvlm}          &   43.89       &     51.43     &   82.49    \\
Phi-3~\cite{phi3}           &   \underline{89.67}      &    \underline{76.75}    &   70.51     \\
\midrule
\rowcolor{gray!15}\multicolumn{4}{c}{\textit{Models Fine-tuned on ChartQA-X}}\\
LLaVA 1.6~\cite{llavanext} &   59.69       &    40.02     &   53.79     \\
\rowcolor{yellow!10}\hspace{2mm} + \textbf{ChartQA-X}  &   65.57 $\scriptstyle\textcolor{darkgreen}{+5.88}$     &   42.85 $\scriptstyle\textcolor{darkgreen}{+2.83}$  &   56.70 $\scriptstyle\textcolor{darkgreen}{+2.91}$   \\
InternVL-2.5~\cite{internvl25}  &   72.57       &   57.57     &   64.64    \\
\rowcolor{yellow!10}\hspace{2mm} + \textbf{ChartQA-X}   &   76.91 $\scriptstyle\textcolor{darkgreen}{+4.34}$  &        68.44 $\scriptstyle\textcolor{darkgreen}{+10.87}$    &   74.09 $\scriptstyle\textcolor{darkgreen}{+9.45}$    \\ 
Qwen2-VL~\cite{qwen2vl}  &   82.57       &     68.13     &   \underline{85.70}    \\
\rowcolor{yellow!10}\hspace{2mm} + \textbf{ChartQA-X}   &   \textbf{97.32} $\scriptstyle\textcolor{darkgreen}{+14.75}$   &        \textbf{78.14} $\scriptstyle\textcolor{darkgreen}{+10.01}$    &   \textbf{95.47} $\scriptstyle\textcolor{darkgreen}{+9.77}$     \\ 

\bottomrule
\end{tabular}
\end{adjustbox}
\caption{Question answering accuracy on three other chart datasets. Best scores are in \textbf{bold}, and second-best scores are \underline{underlined}.}
\label{tab:ablat}
\end{table}

\section{Conclusion}
We introduce ChartQA-X, the largest dataset of visual chart questions paired with detailed explanations.\ Alongside this dataset, we propose a novel carefully-designed explanation generation pipeline that leverages multiple models and evaluates outputs using ROSCOE metrics.
Human evaluation indicates that ChartQA-X explanations are rated comparable to or better than human-written explanations across key dimensions of accuracy, clarity, logic, and overall quality.
Furthermore, models fine-tuned on ChartQA-X significantly outperform existing baselines in generating chart explanations. 
These models also achieve higher accuracy in answering questions on external chart datasets, demonstrating strong generalization and robustness.

\section*{Acknowledgments}
This research was supported by the National Eye Institute (NEI) of the National Institutes of Health (NIH) under award number R01EY034562.\ The content is solely the responsibility of the authors and does not necessarily represent the official views of the NIH. We thank ASU graduate students for contributing human-written chart explanations and ASU Research Computing (RC) for providing computing resources.

{
    \small
    \bibliographystyle{ieeenat_fullname}
    \bibliography{main}
}

\clearpage
\appendix
\section{Appendix}
\label{sec:appendix}


\begin{figure*}[t]
    \centering
    \includegraphics[width=1.\linewidth]{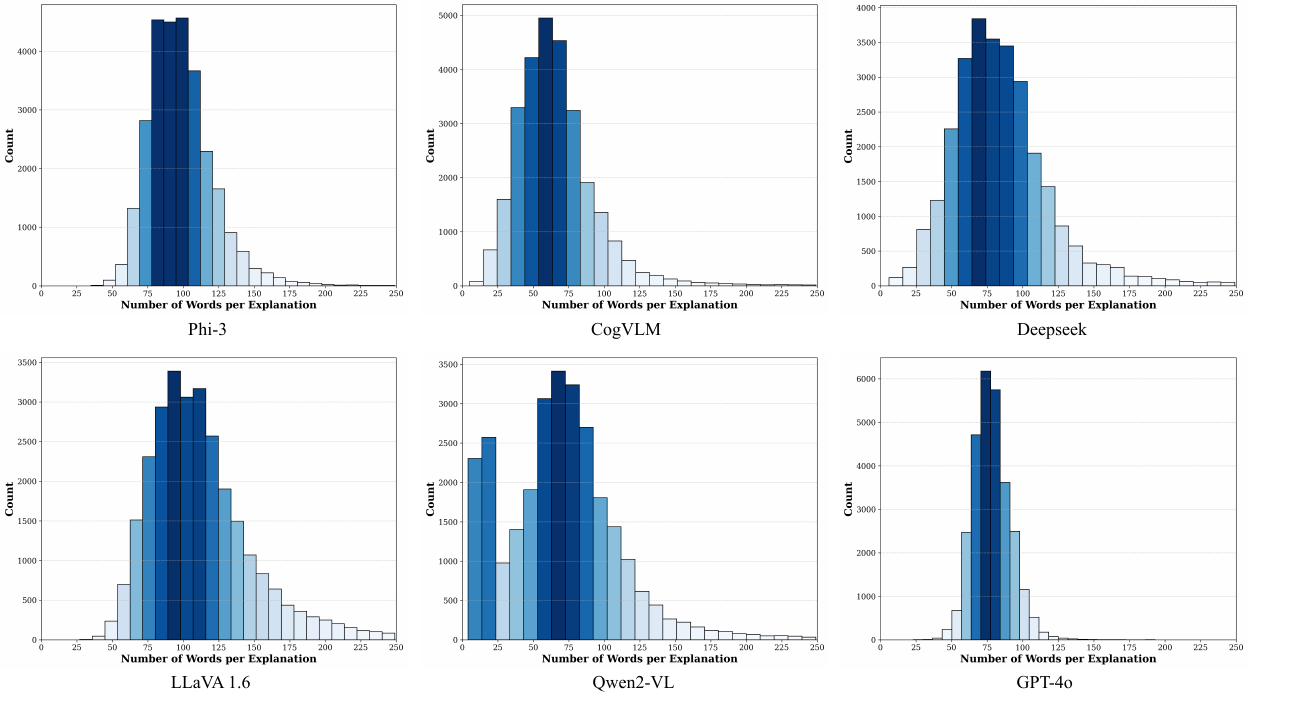}
    \caption{Distribution of length of explanations across different models.}
    \label{fig:worddist}
\end{figure*}

\subsection{Dataset Analysis}

Table \ref{tab:ds_sum} summarizes key statistics of the ChartQA-X dataset, revealing its composition and linguistic characteristics.  The dataset contains 18,317 unique chart images with a varied distribution across chart types: 5,423 horizontal bar charts, 10,158 vertical bar charts, 541 pie charts, and 2,195 line charts.  Accompanying these images are 30,299 questions, averaging 1.54 questions per image.  Explanations for the answers have an average length of 80 words and 401 characters.  The dataset's vocabulary consists of 41,985 unique words, with explanations utilizing an average of 46 unique words each, suggesting a moderate level of lexical diversity.

\begin{table}[h]
    \centering
    \resizebox{\linewidth}{!}{
    \begin{tabular}{ll}
    \toprule
       \textbf{Dataset Statistic}  & \textbf{Value} \\
       \midrule
       Number of Unique Images  &  18,317 \\
       Number of Questions &  30,299 \\
       Number of Horizontal Bar Charts & 5,423 \\
       Number of Vertical Bar Charts & 10,158 \\
       Number of Pie Charts &  541 \\
       Number of Line Charts &  2,195 \\
       Average words per explanation & 80.12 \\
       Average characters per explanation &  401.44 \\
       Number of unique words in the dataset &  41,985 \\
       Number of unique words per explanation &  45.94 \\
       \bottomrule
    \end{tabular}}
    \caption{ChartQA-X Dataset Summary}
    \label{tab:ds_sum}
\end{table}

\subsection{Explanation Lengths}
\label{app:explen}
Figure~\ref{fig:worddist} shows the distribution of explanation lengths (in word count) for six models: Phi-3, CogVLM, Deepseek, LLaVA 1.6, Qwen2-VL, and GPT-4o. These histograms reflect the lengths of explanations across all 30,299 ChartQA samples. We observe notable variation in both the average length and spread across models. Most models generate explanations peaking around 75 words, with word counts mostly falling between 25 and 200 words, and peak frequencies near 4,000 samples. However, GPT-4o differs significantly: it shows a sharper peak and much higher frequency reaching nearly 6,000 samples around the 75-word mark, suggesting more consistent output lengths.
The shape and dispersion of the histograms also vary across models, reflecting underlying differences in model architecture, decoding strategies, and instruction-following behavior. These differences are critical to consider when evaluating the stylistic and structural tendencies of explanations across models.

\subsection{Sensitivity Analysis on ROSCOE Weights}
\label{weight_sens}
To evaluate the robustness of our explanation selection method to different ROSCOE weight configurations, we conduct a sensitivity analysis using eight distinct weighting schemes. These include: \textbf{(1)} our original hand-tuned weights; \textbf{(2)} uniform weights across all nine metrics; \textbf{(3)} random weights sampled from a uniform distribution; \textbf{(4)} random weights from a Gaussian distribution; \textbf{(5)} fluency-focused weights with 0.08 for faithfulness step, 0.07 for faithfulness token, 0.1 for informativeness step and chain, self-consistency and source-consistency, and 0.15 for perplexity step and chain and grammar step; \textbf{(6)} faithfulness-focused weights with 0.30 for faithfulness step, 0.20 for faithfulness token, 0.15 for informativeness step and chain, 0.05 for source-consistency and self-consistency, 0.04 for perplexity step and chain and 0.02 for grammar step; \textbf{(7)} informativeness-focused weights with 0.10 for faithfulness step and token, 0.25 for informativeness step and chain, 0.15 for source-consistency, 0.05 for self-consistency, 0.025 for perplexity step and chain, and 0.05 for grammar step; and \textbf{(8)} a perturbed version of our original weights with small noise added with 0.18 for faithfulness step, 0.17 for faithfulness token, 0.14 for informativeness step, 0.16 for informativeness chain, 0.09 for source-consistency, 0.10 for self-consistency, 0.06 for perplexity step and 0.05 for perplexity chain, and 0.05 for grammar step.

Each weighting scheme is applied separately to compute aggregate ROSCOE scores and select the best explanation among the outputs of all six models for the entire dataset. We then compute the percentage of overlapping best explanation selections between all pairs of weighting sets. The results are shown in Table~\ref{tab:weight_sensitivity}. The first row compares each alternative configuration with our proposed weights.

\begin{table}[h]
\begin{adjustbox}{width=0.5\textwidth,center}
\centering
\begin{tabular}{lccccccc}
\toprule
Compared Sets & Set 2 & Set 3 & Set 4 & Set 5 & Set 6 & Set 7 & Set 8 \\
\midrule
Set 1 (Ours) & 86.87 & 78.55 & 74.90 & 82.21 & 86.39 & 88.55 & \textbf{98.76} \\
Set 2        &       & 82.17 & 76.10 & 94.30 & 74.86 & 81.49 & 87.03 \\
Set 3        &       &       & 74.62 & 82.29 & 77.51 & 74.74 & 78.67 \\
Set 4        &       &       &       & 74.94 & 71.08 & 67.83 & 75.66 \\
Set 5        &       &       &       &       & 79.52 & 78.92 & 82.89 \\
Set 6        &       &       &       &       &       & 79.70 & 84.79 \\
Set 7        &       &       &       &       &       &       & 87.83 \\
\bottomrule
\end{tabular}
\end{adjustbox}
\caption{Agreement (\%) in explanation selection between different ROSCOE weight configurations. Each value indicates the percentage of matching explanation selections. The weight sets are described in Section~\ref{weight_sens}.}
\label{tab:weight_sensitivity}
\end{table}

The analysis reveals that explanation selections are highly consistent across different weight sets. Agreement with our original configuration (Set 1) ranges from 74.90\% (Set 4, random Gaussian) to 98.76\% (Set 8, perturbed). Even randomly assigned weights (Sets 3 and 4) maintain agreement above 74\%. Configurations focused on fluency, faithfulness, and informativeness (Sets 5--7) also perform similarly, with agreement rates over 82\%. We also found that in 54.74\% of the samples, all eight configurations selected the same explanations. These findings indicate that explanation selection using weighted ROSCOE is stable and not overly sensitive to weight configuration.

\begin{table*}[ht]
 \begin{adjustbox}{width=\textwidth,center}
\begin{tabular}{lccccccccc|c}
\toprule
\textbf{Models}    & \textbf{FS}     & \textbf{FT}     & \textbf{IS}     & \textbf{IC}     & \textbf{SRC}    & \textbf{SFC}    & \textbf{PS}     & \textbf{PC}     & \textbf{GS}  & \textbf{AS}   \\ 
\midrule
\rowcolor{gray!15}\multicolumn{11}{c}{\textit{State-of-the-Art Models}}\\
Phi-3~\cite{phi3}             &  0.859  &  0.939  &  0.870  &  0.926  &  0.684  &  0.786  &  0.995  &  0.943  &  0.946  &  0.873 \\
CogVLM~\cite{cogvlm}            &  0.877  &  0.939  &  0.882  &  0.919  &  0.659  &  0.700  &  0.990  &  0.939  &  0.909  &  0.864 \\
Deepseek-VL~\cite{deepseekvl}       &  0.876  &  0.940  &  0.890  &  0.920  &  0.679  &  0.710  &  0.994  &  0.933  &  0.891  &  0.867 \\
GPT-4o~\cite{gpt-4o}            &  0.854  &  0.937  &  0.861  &  0.915  &  0.728  &  0.847  &  0.994  &  0.964  &  0.954  &  0.881 \\
\midrule
\rowcolor{gray!15}\multicolumn{11}{c}{\textit{Models Fine-tuned on ChartQA-X}}\\
InternVL-2.5~\cite{internvl2}      &  0.864  &  0.939  &  0.869  &  0.915  &  0.748  &  0.733  &  0.994  &  0.937  &  0.945  &  0.873 \\
\rowcolor{yellow!10}\hspace{2mm} + \textbf{ChartQA-X}  &  
0.882 $\scriptstyle\textcolor{darkgreen}{+0.018}$  &  
0.940 $\scriptstyle\textcolor{darkgreen}{+0.001}$  &  
\textbf{0.895} $\scriptstyle\textcolor{darkgreen}{+0.026}$  &  
0.926 $\scriptstyle\textcolor{darkgreen}{+0.011}$  &  
\underline{0.800} $\scriptstyle\textcolor{darkgreen}{+0.052}$  &  
0.839 $\scriptstyle\textcolor{darkgreen}{+0.106}$  &  
0.995 $\scriptstyle\textcolor{darkgreen}{+0.001}$  &  
0.966 $\scriptstyle\textcolor{darkgreen}{+0.029}$  &  
0.949 $\scriptstyle\textcolor{darkgreen}{+0.004}$  &  
0.900 $\scriptstyle\textcolor{darkgreen}{+0.013}$ \\
LLaVA 1.6~\cite{llavanext}         &  0.889  &  0.949  &  0.880  &  0.922  &  0.529  &  0.847  &  0.989  &  0.950  &  0.931  &  0.872 \\
\rowcolor{yellow!10}\hspace{2mm} + \textbf{ChartQA-X}  &  
\textbf{0.925} $\scriptstyle\textcolor{darkgreen}{+0.036}$  &  
\underline{0.952} $\scriptstyle\textcolor{darkgreen}{+0.003}$  &  
\textbf{0.895} $\scriptstyle\textcolor{darkgreen}{+0.015}$  &  
\underline{0.930} $\scriptstyle\textcolor{darkgreen}{+0.008}$  &  
0.728 $\scriptstyle\textcolor{darkgreen}{+0.199}$  &  
\textbf{0.984} $\scriptstyle\textcolor{darkgreen}{+0.137}$  &  
\textbf{0.997} $\scriptstyle\textcolor{darkgreen}{+0.008}$  &  
\underline{0.977} $\scriptstyle\textcolor{darkgreen}{+0.027}$  &  
\underline{0.955} $\scriptstyle\textcolor{darkgreen}{+0.024}$  &  
0.919 $\scriptstyle\textcolor{darkgreen}{+0.023}$ \\
Qwen2-VL~\cite{qwen2vl}      &  0.877  &  0.943  &  0.876  &  0.922  &  0.750  &  0.879  &  0.995  &  0.964  &  0.946  &  0.895 \\
\rowcolor{yellow!10}\hspace{2mm} + \textbf{ChartQA-X}  &  
\underline{0.905} $\scriptstyle\textcolor{darkgreen}{+0.028}$  &  
\textbf{0.953} $\scriptstyle\textcolor{darkgreen}{+0.010}$  &  
\underline{0.892} $\scriptstyle\textcolor{darkgreen}{+0.016}$  &  
\textbf{0.963} $\scriptstyle\textcolor{darkgreen}{+0.041}$  &  
\textbf{0.901} $\scriptstyle\textcolor{darkgreen}{+0.151}$  &  
\underline{0.917} $\scriptstyle\textcolor{darkgreen}{+0.039}$  &  
\textbf{0.997} $\scriptstyle\textcolor{darkgreen}{+0.002}$  &  
\textbf{0.988} $\scriptstyle\textcolor{darkgreen}{+0.024}$  &  
\textbf{0.957} $\scriptstyle\textcolor{darkgreen}{+0.011}$  &  
0.931 $\scriptstyle\textcolor{darkgreen}{+0.049}$ \\
\bottomrule
\end{tabular}
\end{adjustbox}
\caption{ROSCOE scores on the ChartQA-X test set with data table included as input. Best scores are in \textbf{bold}, and second-best scores are \underline{underlined}. FS: Faithfulness Step, FT: Faithfulness Token, IS: Informativeness Step, IC: Informativeness Chain, SRC: Source-Consistency, SFC: Self-Consistency, PS: Perplexity Step,  PC: Perplexity Chain, GS: Grammar Step, and AS: Aggregate Score.}
\label{tab:bench_roscoe_wdt}
\end{table*}

\subsection{ROSCOE Scores}
\label{app:results}

Tables~\ref{tab:bench_roscoe_wdt}, \ref{tab:dvqa_roscoe},  \ref{tab:plotqa_roscoe}, and \ref{tab:figureqa_roscoe} present ROSCOE scores for explanations generated by various VLMs on ChartQA-X with data table, DVQA, PlotQA, and FigureQA, respectively.

\subsection{Qualitative Analysis}
\label{app:examples}
\begin{figure*}[t]
    \centering
    \includegraphics[width=1.\linewidth]{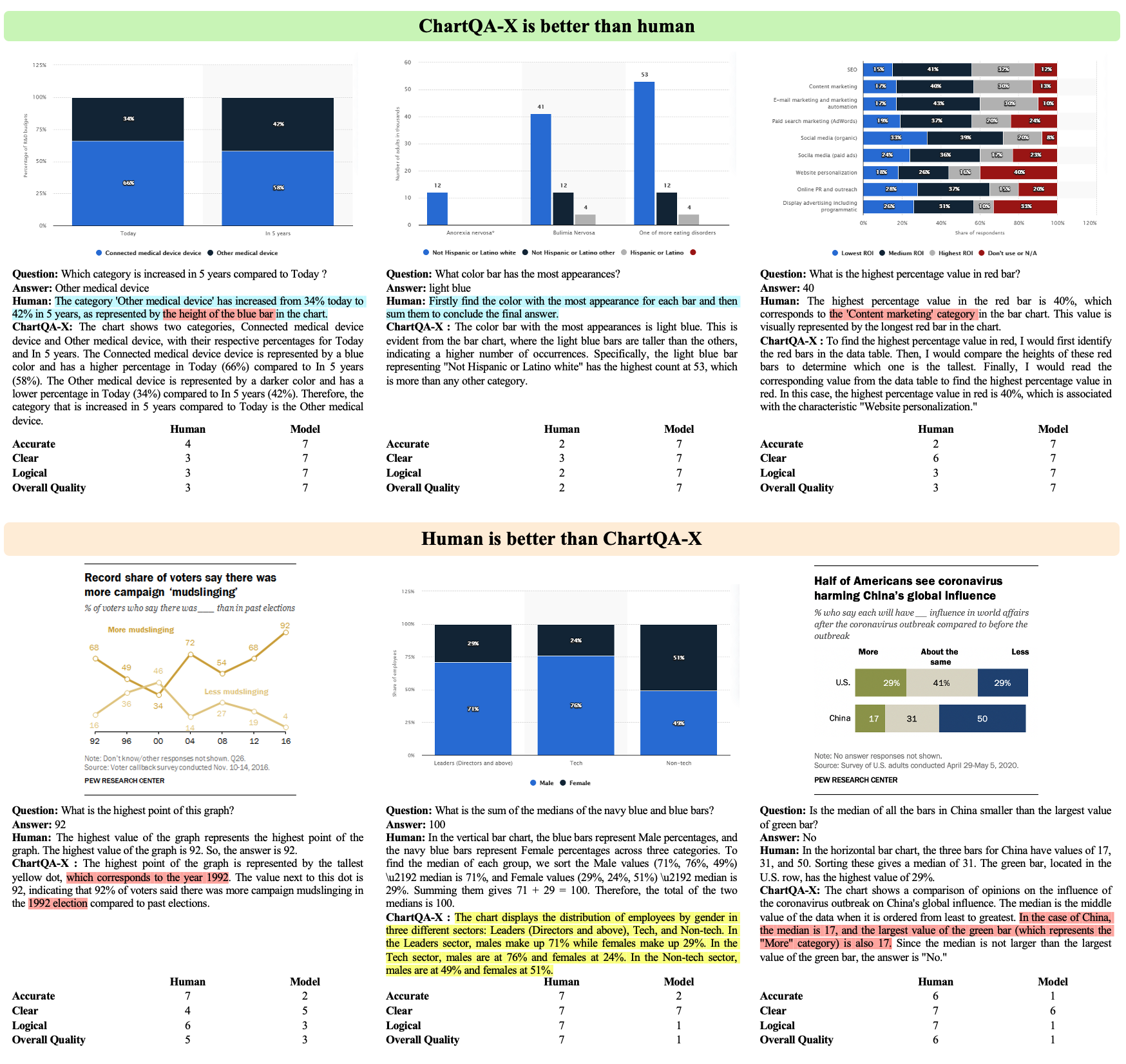}
    \caption{Representative qualitative comparisons between human-written and ChartQA-X explanations from the MTurk study. Each example presents the chart question, the correct answer, and paired explanations from both human and our dataset. Explanations are accompanied by Likert-scale ratings (1–7) across four evaluation criteria: Accurate, Clear, Logical, and Overall Quality. The selected examples showcase a range of scenarios in which the ChartQA-X explanation either outperforms (top row) or underperforms (bottom row) the human-written ones. Brief annotations at the top of each chart describe the rationale behind each set of user ratings. Red, yellow, and blue highlights indicate explanation segments that are incorrect, irrelevant to the answer, or missing step-by-step reasoning, respectively.}
    \label{fig:qualitative_human}
\end{figure*}

Figure~\ref{fig:qualitative_human} shows six examples from our MTurk study in which ChartQA-X explanations either outperform or underperform those written by humans.

\begin{table*}[ht]
\centering
\begin{adjustbox}{width=\textwidth,center}
\begin{tabular}{lcccccccccc}
\hline
\textbf{Models}                           
& \textbf{FS} & \textbf{FT} & \textbf{IS} & \textbf{IC} & \textbf{SRC} & \textbf{SFC} & \textbf{PS} & \textbf{PC} & \textbf{GS} & \textbf{AS}  \\ \midrule
\rowcolor{gray!15}\multicolumn{11}{c}{\textit{State-of-the-Art Models}}\\ 
Deepseek-VL~\cite{deepseekvl}  &   0.889    &  0.942  &  0.885  &  0.913   &   0.657  &  0.777  &  0.994  & 0.945  &  0.914  &  0.875 \\
CogVLM~\cite{cogvlm}   &  0.871  &  0.939 & 0.879  &  0.932  &  0.760  &  0.854 &  0.995 & 0.969  & 0.911  & 0.892  \\
Phi-3~\cite{phi3}    &  0.879  & 0.939  &  0.883  &  0.932  &  0.690  &  0.736   &  \underline{0.996}  &  0.948  &  0.902 & 0.874 \\
\midrule
\rowcolor{gray!15}\multicolumn{11}{c}{\textit{Models Fine-tuned on ChartQA-X}}\\
LLaVA 1.6~\cite{llavanext} &  0.881  &  0.938  &  0.890  & 0.932  & 0.507   &  0.759  &  \textbf{0.997}  & 0.949  &  0.885  & 0.858  \\
\rowcolor{yellow!10}\hspace{2mm} + \textbf{ChartQA-X} 
& \underline{0.922} $\scriptstyle\textcolor{darkgreen}{+0.041}$  
& \underline{0.954} $\scriptstyle\textcolor{darkgreen}{+0.016}$  
& \underline{0.899} $\scriptstyle\textcolor{darkgreen}{+0.009}$  
& 0.930 $\scriptstyle\textcolor{darkred}{-0.002}$  
& 0.715 $\scriptstyle\textcolor{darkgreen}{+0.208}$  
& 0.941 $\scriptstyle\textcolor{darkgreen}{+0.182}$  
& \textbf{0.997} $\scriptstyle\textcolor{darkgreen}{+0.000}$  
& \underline{0.988} $\scriptstyle\textcolor{darkgreen}{+0.039}$  
& 0.902 $\scriptstyle\textcolor{darkgreen}{+0.017}$  
& 0.912 $\scriptstyle\textcolor{darkgreen}{+0.054}$ \\
InternVL-2.5~\cite{internvl25}  & 0.881  &  0.939  & 0.890  & 0.928  & 0.688   & 0.753   &  \underline{0.996} & 0.946 & 0.909  & 0.876 \\
\rowcolor{yellow!10}\hspace{2mm} + \textbf{ChartQA-X}    
& 0.907 $\scriptstyle\textcolor{darkgreen}{+0.026}$  
& \textbf{0.955} $\scriptstyle\textcolor{darkgreen}{+0.016}$  
& 0.884 $\scriptstyle\textcolor{darkred}{-0.006}$  
& \underline{0.940} $\scriptstyle\textcolor{darkgreen}{+0.012}$  
& \textbf{0.810} $\scriptstyle\textcolor{darkgreen}{+0.122}$  
& \underline{0.950} $\scriptstyle\textcolor{darkgreen}{+0.197}$  
& \underline{0.996} $\scriptstyle\textcolor{darkgreen}{+0.000}$  
& 0.980 $\scriptstyle\textcolor{darkgreen}{+0.034}$  
& \textbf{0.919} $\scriptstyle\textcolor{darkgreen}{+0.010}$  
& \underline{0.918} $\scriptstyle\textcolor{darkgreen}{+0.042}$ \\
Qwen2-VL~\cite{qwen2vl} & 0.887 & 0.942 & 0.878 & 0.929 & 0.695 & 0.870 & 0.995 & 0.963 & \underline{0.915} & 0.890 \\
\rowcolor{yellow!10}\hspace{2mm} + \textbf{ChartQA-X}   
& \textbf{0.926} $\scriptstyle\textcolor{darkgreen}{+0.039}$  
& \underline{0.954} $\scriptstyle\textcolor{darkgreen}{+0.012}$  
& \textbf{0.902} $\scriptstyle\textcolor{darkgreen}{+0.024}$  
& \textbf{0.967} $\scriptstyle\textcolor{darkgreen}{+0.038}$  
& \underline{0.789} $\scriptstyle\textcolor{darkgreen}{+0.094}$  
& \textbf{0.990} $\scriptstyle\textcolor{darkgreen}{+0.120}$  
& 0.995 $\scriptstyle\textcolor{darkgreen}{+0.000}$  
& \textbf{0.991} $\scriptstyle\textcolor{darkgreen}{+0.028}$  
& 0.834 $\scriptstyle\textcolor{darkred}{-0.081}$  
& \textbf{0.927} $\scriptstyle\textcolor{darkgreen}{+0.037}$ \\ \hline
\end{tabular}
 \end{adjustbox}
\caption{ROSCOE scores on DVQA. Best scores are in \textbf{bold}, and second-best scores are \underline{underlined}. FS: Faithfulness Step, FT: Faithfulness Token, IS: Informativeness Step, IC: Informativeness Chain, SRC: Source-Consistency, SFC: Self-Consistency, PS: Perplexity Step,  PC: Perplexity Chain, GS: Grammar Step, and AS: Aggregate Score.}
\label{tab:dvqa_roscoe}
\end{table*}

\clearpage

\begin{table*}[ht]
\centering
\begin{adjustbox}{width=\textwidth,center}
\begin{tabular}{lcccccccccc}
\hline
\textbf{Models}                           
& \textbf{FS} & \textbf{FT} & \textbf{IS} & \textbf{IC} & \textbf{SRC} & \textbf{SFC} & \textbf{PS} & \textbf{PC} & \textbf{GS} & \textbf{AS}  \\ \midrule
\rowcolor{gray!15}\multicolumn{11}{c}{\textit{State-of-the-Art Models}}\\
Deepseek-VL~\cite{deepseekvl}  &  0.917  &  0.956 & \underline{0.902}  &  0.955  & 0.700   & 0.827  & 0.990  & 0.946  &  0.937 &  0.902 \\
CogVLM~\cite{cogvlm}   &  0.885  & 0.949  & 0.882  & 0.959  &  0.165  &  0.398  & \underline{0.996}  & 0.907  & \underline{0.946}  & 0.794  \\
Phi-3~\cite{phi3}  & 0.900 & 0.946 & 0.884 & 0.940 & 0.545 & 0.688 & 0.995 & 0.940 & 0.915 & 0.861  \\
\midrule
\rowcolor{gray!15}\multicolumn{11}{c}{\textit{Models Fine-tuned on ChartQA-X}}\\
LLaVA 1.6~\cite{llavanext} & 0.910 & 0.952 & 0.889 & 0.948 & 0.615 & 0.745 & \textbf{0.997} & 0.947 & 0.922 & 0.879 \\
\rowcolor{yellow!10}\hspace{2mm} + \textbf{ChartQA-X}    
& \underline{0.925} $\scriptstyle\textcolor{darkgreen}{+0.015}$ 
& 0.957 $\scriptstyle\textcolor{darkgreen}{+0.005}$ 
& 0.894 $\scriptstyle\textcolor{darkgreen}{+0.005}$ 
& 0.952 $\scriptstyle\textcolor{darkgreen}{+0.004}$ 
& \textbf{0.790} $\scriptstyle\textcolor{darkgreen}{+0.175}$ 
& \underline{0.865} $\scriptstyle\textcolor{darkgreen}{+0.120}$ 
& \textbf{0.997} $\scriptstyle\textcolor{darkgreen}{+0.000}$ 
& \underline{0.961} $\scriptstyle\textcolor{darkgreen}{+0.014}$ 
& 0.930 $\scriptstyle\textcolor{darkgreen}{+0.008}$ 
& \underline{0.915} $\scriptstyle\textcolor{darkgreen}{+0.036}$ \\
InternVL-2.5~\cite{internvl25}  
&  0.902 & 0.945 & 0.882 & 0.943 & 0.582 & 0.719 & 0.995 & 0.939 & 0.920 & 0.869 \\
\rowcolor{yellow!10}\hspace{2mm} + \textbf{ChartQA-X} 
& 0.920 $\scriptstyle\textcolor{darkgreen}{+0.018}$ 
& \textbf{0.960} $\scriptstyle\textcolor{darkgreen}{+0.015}$ 
& 0.886 $\scriptstyle\textcolor{darkgreen}{+0.004}$ 
& \underline{0.963} $\scriptstyle\textcolor{darkgreen}{+0.020}$ 
& 0.725 $\scriptstyle\textcolor{darkgreen}{+0.143}$ 
& 0.825 $\scriptstyle\textcolor{darkgreen}{+0.106}$ 
& \textbf{0.997} $\scriptstyle\textcolor{darkgreen}{+0.002}$ 
& 0.965 $\scriptstyle\textcolor{darkgreen}{+0.026}$ 
& \textbf{0.950} $\scriptstyle\textcolor{darkgreen}{+0.030}$ 
& 0.906 $\scriptstyle\textcolor{darkgreen}{+0.037}$ \\
Qwen2-VL~\cite{qwen2vl} 
& 0.908 & 0.948 & 0.888 & 0.945 & 0.700 & 0.800 & 0.994 & 0.950 & 0.924 & 0.892 \\
\rowcolor{yellow!10}\hspace{2mm} + \textbf{ChartQA-X}   
& \textbf{0.930} $\scriptstyle\textcolor{darkgreen}{+0.022}$  
& \underline{0.959} $\scriptstyle\textcolor{darkgreen}{+0.011}$  
& \textbf{0.906} $\scriptstyle\textcolor{darkgreen}{+0.018}$  
& \textbf{0.969} $\scriptstyle\textcolor{darkgreen}{+0.024}$  
& \underline{0.771} $\scriptstyle\textcolor{darkgreen}{+0.071}$  
& \textbf{0.962} $\scriptstyle\textcolor{darkgreen}{+0.162}$  
& 0.990 $\scriptstyle\textcolor{darkred}{-0.004}$  
& \textbf{0.983} $\scriptstyle\textcolor{darkgreen}{+0.033}$  
& 0.907 $\scriptstyle\textcolor{darkred}{-0.017}$  
& \textbf{0.928} $\scriptstyle\textcolor{darkgreen}{+0.036}$ \\ \hline
\end{tabular}
\end{adjustbox}
\caption{ROSCOE scores on PlotQA. Best scores are in \textbf{bold}, and second-best scores are \underline{underlined}. FS: Faithfulness Step, FT: Faithfulness Token, IS: Informativeness Step, IC: Informativeness Chain, SRC: Source-Consistency, SFC: Self-Consistency, PS: Perplexity Step,  PC: Perplexity Chain, GS: Grammar Step, and AS: Aggregate Score.}
\label{tab:plotqa_roscoe}
\end{table*}

\begin{table*}[ht]
\centering
\begin{adjustbox}{width=\textwidth,center}
\begin{tabular}{lcccccccccc}
\hline
\textbf{Models}
& \textbf{FS} & \textbf{FT} & \textbf{IS} & \textbf{IC} & \textbf{SRC} & \textbf{SFC} & \textbf{PS} & \textbf{PC} & \textbf{GS} & \textbf{AS} \\ \midrule
\rowcolor{gray!15}\multicolumn{11}{c}{\textit{State-of-the-Art Models}}\\
Deepseek-VL~\cite{deepseekvl}  
& 0.921 & 0.933 & \underline{0.880} & 0.892 & \textbf{0.480} & \underline{0.989} & 0.989 & \underline{0.985} & \textbf{0.961} & \underline{0.882} \\
CogVLM~\cite{cogvlm}   
& 0.921 & 0.935 & 0.878 & 0.910 & 0.448 & 0.940 & 0.995 & 0.980 & 0.944 & 0.877 \\
Phi-3~\cite{phi3}    
& 0.849 & 0.932 & 0.858 & \textbf{0.947} & 0.328 & 0.442 & 0.995 & 0.938 & 0.913 & 0.799 \\
\midrule
\rowcolor{gray!15}\multicolumn{11}{c}{\textit{Models Fine-tuned on ChartQA-X}}\\
LLaVA 1.6~\cite{llavanext} 
& 0.918 & 0.932 & 0.872 & 0.901 & 0.300 & 0.406 & 0.995 & 0.940 & 0.935 & 0.803 \\
\rowcolor{yellow!10}\hspace{2mm} + \textbf{ChartQA-X}    
& \underline{0.923} $\scriptstyle\textcolor{darkgreen}{+0.005}$ 
& \textbf{0.938} $\scriptstyle\textcolor{darkgreen}{+0.006}$ 
& 0.875 $\scriptstyle\textcolor{darkgreen}{+0.003}$ 
& 0.914 $\scriptstyle\textcolor{darkgreen}{+0.013}$ 
& 0.445 $\scriptstyle\textcolor{darkgreen}{+0.145}$ 
& 0.920 $\scriptstyle\textcolor{darkgreen}{+0.514}$ 
& \underline{0.996} $\scriptstyle\textcolor{darkgreen}{+0.001}$ 
& 0.981 $\scriptstyle\textcolor{darkgreen}{+0.041}$ 
& 0.945 $\scriptstyle\textcolor{darkgreen}{+0.010}$ 
& 0.876 $\scriptstyle\textcolor{darkgreen}{+0.073}$ \\
InternVL-2.5~\cite{internvl25}  
& 0.912 & 0.929 & 0.870 & 0.900 & 0.390 & 0.925 & \underline{0.996} & 0.980 & 0.940 & 0.865 \\
\rowcolor{yellow!10}\hspace{2mm} + \textbf{ChartQA-X} 
& 0.920 $\scriptstyle\textcolor{darkgreen}{+0.008}$ 
& 0.926 $\scriptstyle\textcolor{darkred}{-0.003}$ 
& \textbf{0.882} $\scriptstyle\textcolor{darkgreen}{+0.012}$ 
& 0.894 $\scriptstyle\textcolor{darkred}{-0.006}$ 
& 0.461 $\scriptstyle\textcolor{darkgreen}{+0.071}$ 
& \textbf{0.993} $\scriptstyle\textcolor{darkgreen}{+0.068}$ 
& 0.990 $\scriptstyle\textcolor{darkred}{-0.006}$ 
& 0.983 $\scriptstyle\textcolor{darkgreen}{+0.003}$ 
& \underline{0.948} $\scriptstyle\textcolor{darkgreen}{+0.008}$ 
& 0.881 $\scriptstyle\textcolor{darkgreen}{+0.016}$ \\
Qwen2-VL~\cite{qwen2vl} 
& 0.915 & 0.930 & 0.871 & 0.898 & 0.422 & 0.942 & 0.993 & 0.974 & 0.938 & 0.870 \\
\rowcolor{yellow!10}\hspace{2mm} + \textbf{ChartQA-X}   
& \textbf{0.925} $\scriptstyle\textcolor{darkgreen}{+0.010}$ 
& \underline{0.937} $\scriptstyle\textcolor{darkgreen}{+0.007}$ 
& \textbf{0.882} $\scriptstyle\textcolor{darkgreen}{+0.011}$ 
& \underline{0.915} $\scriptstyle\textcolor{darkgreen}{+0.017}$ 
& \underline{0.468} $\scriptstyle\textcolor{darkgreen}{+0.046}$ 
& 0.948 $\scriptstyle\textcolor{darkgreen}{+0.006}$ 
& \textbf{0.997} $\scriptstyle\textcolor{darkgreen}{+0.004}$ 
& \textbf{0.986} $\scriptstyle\textcolor{darkgreen}{+0.012}$ 
& \underline{0.948} $\scriptstyle\textcolor{darkgreen}{+0.010}$ 
& \textbf{0.883} $\scriptstyle\textcolor{darkgreen}{+0.013}$ \\ \hline
\end{tabular}
\end{adjustbox}
\caption{ROSCOE scores on FigureQA. Best scores are in \textbf{bold}, and second-best scores are \underline{underlined}. FS: Faithfulness Step, FT: Faithfulness Token, IS: Informativeness Step, IC: Informativeness Chain, SRC: Source-Consistency, SFC: Self-Consistency, PS: Perplexity Step,  PC: Perplexity Chain, GS: Grammar Step, and AS: Aggregate Score.}
\label{tab:figureqa_roscoe}
\end{table*}

\end{document}